\documentclass[10pt,twocolumn,letterpaper]{article}

\usepackage{iccv}
\usepackage{times}
\usepackage{epsfig}
\usepackage{graphicx}
\usepackage{amsmath}
\usepackage{amssymb}
\usepackage{bbm}
\usepackage{tablefootnote}
\usepackage{booktabs}
\usepackage{subcaption}
\usepackage{microtype}
\usepackage{paralist}

\usepackage{flushend}
\usepackage[accsupp]{axessibility}  

\usepackage[pagebackref=true,breaklinks=true,letterpaper=true,colorlinks,bookmarks=false]{hyperref}

\newcommand{\rYC}{YR10}
\newcommand{\rMSRVTT}{MR10}
\newcommand{\rCrossTask}{CTR}

\newcommand{\tablestyle}[2]{\setlength{\tabcolsep}{#1}\renewcommand{\arraystretch}{#2}\centering\footnotesize}

\usepackage{mathtools,xparse}

\iccvfinalcopy 
\DeclarePairedDelimiter{\norm}{\lVert}{\rVert}
\def\iccvPaperID{2965} 

\ificcvfinal\fi

\begin{document}

\title{Multimodal Clustering Networks for \\ Self-supervised  Learning  from Unlabeled Videos}

\author{%
    Brian Chen $^1$  \quad
    Andrew Rouditchenko$^2$  \quad
    Kevin Duarte $^3$ \quad
    Hilde Kuehne$^{4,6}$ \quad
    Samuel Thomas$^{5,6}$  \vspace{1mm} \\
    Angie Boggust$^2$ \quad
    Rameswar Panda$^{5,6}$  \quad
    Brian Kingsbury$^{5,6}$ \quad
    Rogerio Feris$^{5,6}$ \vspace{1mm} \\
    David Harwath$^7$  \quad 
    James Glass$^2$ \quad 
    Michael Picheny$^{8}$ \quad 
    Shih-Fu Chang$^{1}$ 
    \vspace{1mm} \\
    \small{$^1$Columbia University, $^2$MIT CSAIL, $^3$University of Central Florida,  $^4$Goethe University Frankfurt},  \\
    \small{$^5$IBM Research AI, $^6$MIT-IBM Watson AI Lab, $^7$UT Austin, $^8$NYU-Courant CS \& CDS,} \\
    \small{
    \texttt{\{bc2754,sc250\}@columbia.edu},\texttt{\{roudi,aboggust,glass\}@mit.edu},\texttt{kevin\_duarte@knights.ucf.edu}} \\ \small{\texttt{\{kuehne,rpanda\}@ibm.com},\texttt{\{sthomas,rsferis,bedk\}@us.ibm.com}, } 
    \small{\texttt{harwath@cs.utexas.edu}, \texttt{map22@nyu.edu}   }  
}

\maketitle

\begin{abstract}
Multimodal self-supervised learning is getting more and more attention as it allows not only to train large networks without human supervision but also to search and retrieve data across various modalities. In this context, this paper proposes a framework that, starting from a pre-trained backbone, learns a common multimodal embedding space that, in addition to sharing representations across different modalities, enforces a grouping of semantically similar instances. To this end, we extend the concept of instance-level contrastive learning with a multimodal clustering step in the training pipeline to capture semantic similarities across modalities. The resulting embedding space enables retrieval of samples across all modalities, even from unseen datasets and different domains.
To evaluate our approach, we train our model on the HowTo100M dataset and evaluate its zero-shot retrieval capabilities in two challenging domains, namely text-to-video retrieval, and temporal action localization, showing state-of-the-art results on four different datasets.
\end{abstract}

\section{Introduction}


To robustly learn visual events and concepts, humans seldom rely on visual inputs alone. Instead, a rich multimodal environment is utilized for understanding by combining multiple sensory signals along with various language representations. 
Many recent techniques have attempted to mimic this paradigm to train efficient computer vision models, especially those that learn from videos where multiple modalities are naturally present~\cite{alayrac2020self,alwassel2020self, piergiovanni2020evolving}. 


 \begin{figure}[bpt]
 \resizebox{\columnwidth}{!}{
  \includegraphics[width=\columnwidth]{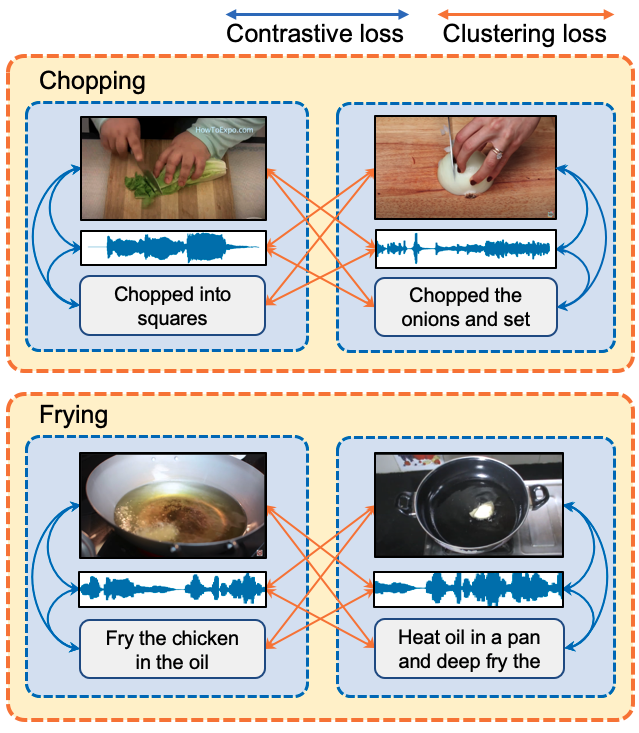}}
\caption{The Multimodal Clustering Network (MCN) combines a contrastive loss that learns feature representations to be close across different modalities such as video, audio, and text (blue box), with a clustering loss that draws instances that are semantically related together, e.g., scenes depicting the same semantic concept (e.g., chopping or frying) from different videos or different clips. (yellow box).}
\label{fig:open}
\vspace{-6mm}
\end{figure}

Learning on multimodal video data has both benefits and challenges. It is beneficial that each video instance has information available in multiple modalities. Textual information corresponding to the spoken narrations in the video, for example, provides a valuable language modality in addition to the visual and audio modalities \cite{aytar2017see,harwath2018jointly,kaiser2017one}. 
In this work, we focus on the problem of learning a joint embedding space across multiple modalities.
Given that the features from different modalities are often not comparable, the goal is to learn the projections into a common space where features from different domains but with similar content are close to each other to allow for a direct retrieval across modalities. 
However, creating an effective joint multimodal embedding space is not easy. First, each of those modalities is different, \ie with respect to its source, how it is sampled and processed, and its resulting feature representation. 
Additionally, in real-world data, the supervision available to learn these projections from each of the modalities is unfortunately weak, as \eg audio sequences can be misaligned to their visual representations and corresponding narration might or might not be present in the same time interval \cite{alwassel2020self,miech2020end}.

To deal with multimodal data of this nature, several recent approaches use a contrastive loss~ \cite{gutmann2010noise,hadsell2006dimensionality} to learn e.g. feature representations in a joint embedding space. The goal is to bring samples drawn from the same temporal instance closer to each other while keeping samples from different times apart. Recent works~\cite{alayrac2020self,miech2020end} show that such training is useful for pretraining models on large-scale data without additional supervision and that the resulting models achieve competitive performance on several tasks, \eg in action classification when fine-tuned on various datasets. 
One problem arising from the contrastive loss is that this criterion does not consider the samples' semantic structure and similarity at different times: two samples are treated as a negative pair as long as they occur at different times regardless of their semantic similarity. This can have a considerable adverse impact on the learned representation. In a different formulation for learning representations, instead of comparing individual instances, clusters of instances are first created using a certain clustering algorithm~\cite{alwassel2020self,asano2020labelling,caron2020unsupervised,li2020prototypical}. This approach encourages samples semantically similar to each other (namely, samples in the same cluster) to be close in the embedding space. 
However, if we cluster features from multi-modalities, those clusters would likely emerge only within the modalities separately, clustering audio instances with audio instances, visuals to visuals \etc. Therefore, a mechanism that pulls the instances from different modalities together is crucial to cluster features from different modalities in a joint space. This leads to our proposed method that treats these two approaches as reciprocal information.


We present a multimodal learning framework that learns joint representations by training cross-modal projection heads from the visual, audio, and language modalities and accounts for the semantic similarity of embedding using a large corpus of naturally narrated videos. The proposed \textit{Multimodal Clustering Network} (MCN) adopts a novel architecture to combine promising ideas from both representation learning paradigms described earlier: learning via the contrastive loss at the instance level and the semantic consistency at the cluster level. 
As another novel feature of our approach, we explore joint clusters using multimodal representations instead of clusters using separate modalities. The result features allow us to do retrieval across different modalities in linear time. Figure \ref{fig:open} provides a high-level overview of our approach. 


To evaluate our proposed method, we address the challenging problem of zero-shot learning in two contexts: multimodal video retrieval and multimodal temporal action localization. 
We train our system on the HowTo100M dataset \cite{miech2019howto100m} and evaluate its retrieval capabilities on the YouCook2 \cite{zhou18towards} and MSR-VTT \cite{xu16msrvtt} dataset and its temporal action localization on the task of action detection on the CrossTask \cite{zhukov2019crosstask} dataset and on the task of temporal action segmentation on the Mining YouTube \cite{kuehne2019mining} dataset.
Using only features from pretrained backbones, MCN significantly outperforms the best text-to-video retrieval baseline over absolute $3\%$ in recall and outperforms the temporal action localization baseline over $3.1\%$ in recall, both in zero-shot settings.



\textbf{Contributions.} The contributions of this work are threefold:
\begin{inparaenum}[(i)]
\item We propose a novel method by combining the benefits of contrastive loss and clustering loss for multimodal joint space learning. Unlike prior works that create clusters using separate modalities, our method shows the important benefits of using multimodal joint clusters.
\item We show that the proposed model can learn across three modalities (video, audio, text) in a joint space. 
\item We demonstrate significant performance gains on multiple downstream tasks in the zero-shot setting. These results show that the learned common space representations can improve state-of-the-art results without any additional training on the target datasets.
\end{inparaenum}

\section{Related Work}


\noindent \textbf{Learning from Multimodal Data. }
Instead of collecting new annotated datasets 
~\cite{carreira2017quo,imagenet} for building various state-of-the-art visual recognition models, current approaches leverage large amounts of videos available on multiple social media platforms. 
When specific language resources like automatically generated speech recognition captions are available in narrated video datasets such as How2~\cite{sanabria18how2} or HowTo100M~\cite{miech2019howto100m}, an appropriate proxy task that leverages these resources is instead used. 
Such visual caption pairs have been widely used in self-supervised models in vision and language tasks recently \cite{amrani2020noise,dong2021dual,gabeur2020multi,lei2021less,luo2020univilm,patrick2020support,sun2019videobert,zhu2020actbert}.
In other approaches like~\cite{alwassel2020self,asano2019self,boggust2019grounding, harwath2018jointly, liu2019use, rouditchenko2020avlnet}, the need for these language transcripts is avoided by using just the corresponding raw speech signal. 
More recently, models that trained from scratch from the narrated video along with generated speech captions have also been successfully developed \cite{miech2020end}. 
The three modalities naturally present in videos, the visual, audio, and language streams, are further integrated via a multimodal variant of this learning framework in \cite{alayrac2020self}. Unlike these works, our goal in this paper is to learn a joint embedding in three modalitites for zero-shot multimodal downstream tasks where we create an embedding space which the features across different modalitites are directly comparable.


\noindent \textbf{Contrastive Learning. }  
A technique central to several state-of-the-art self-supervised representation learning approaches for images is instance-wise contrastive learning \cite{chen2020simple,he2020momentum}. In this paradigm, a model is trained to place samples extracted from the same instance, e.g., transforms or crops of an image, close to each other while pushing samples from different instances further apart. 
Given its similarity to noise contrastive estimation (NCE), where two samples are treated as a negative pair as long as they are drawn from different time segments, in MIL-NCE \cite{miech2020end}, the benefits of both multiple instance learning and NCE are combined. An advantage of this approach is that it now allows for compensation of misalignments inherently found in videos and corresponding text captions. 
One inherent drawback of the instance-wise contrastive learning described above is that it is agnostic to the inherent semantic similarity between the samples when positive and negative pairs are constructed. In our work, we alleviate this problem by relaxing the instance level similarity across modalities to semantic level similarity by introducing a clustering component that learns semantic similarity among multimodal instances within the batch.



 \begin{figure}[t]
 \resizebox{\columnwidth}{!}{
 \centering
  \includegraphics[width=\columnwidth]{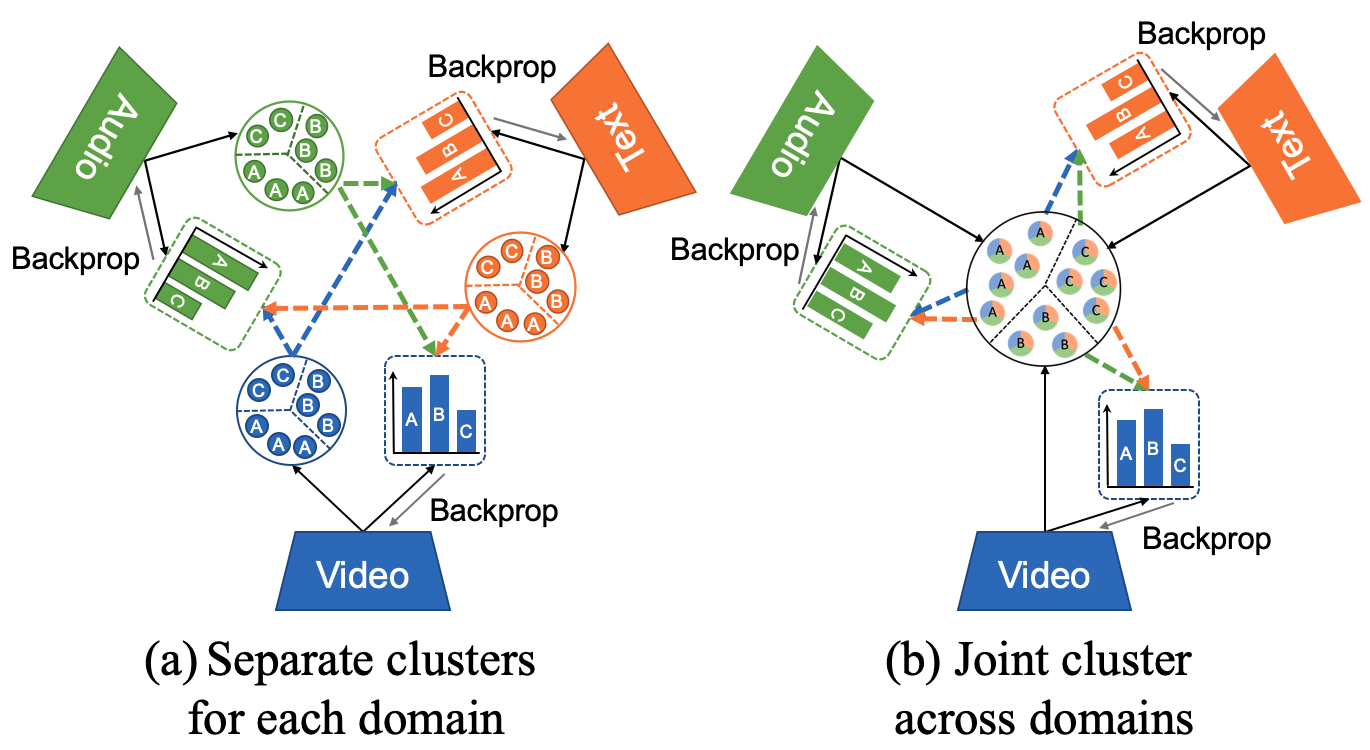}}
\caption{\textbf{Cross-domain Clustering vs.\ Joint Clustering.} (a) Previous methods such as XDC perform clustering at separate spaces and use pseudo-labels as supervision to other domains. (b) Our method performs clustering across features from different modalities in the joint space to learn multimodal clusters. Best viewed in color.}
\vspace{-4mm}
\label{fig:compare}
\end{figure}

\noindent \textbf{Deep Unsupervised Clustering. } 
Given the high cost of computing all pairwise comparisons in a large dataset, instead of applying the contrastive learning paradigm discussed above on each individual instance, a more practical solution is to discriminate between groups of instances during training. This is done by first pre-training a model to derive suitable feature representations of the data in a simple cascaded approach. Keeping the representations fixed, a clustering algorithm is then used to group instances before the weights of the model are updated using the derived class assignments as supervision \cite{caron2018deep,yan2020clusterfit}. In contrast, instead of keeping the clustering step independent of the representation learning phase, more recent techniques jointly learn visual embeddings and cluster assignments \cite{asano2020labelling,asano2019self,caron2020unsupervised,van2020scan}.
While both these approaches can produce interpretable clustering results that benefit downstream tasks by integrating global information across the entire dataset, running a clustering algorithm over a large data set slows down training. However, this issue can be addressed by performing the clustering in an online fashion \cite{caron2020unsupervised}. These online models simultaneously learn to cluster and represent image data. To improve the performance of clustering, it is, however, also essential to leverage the correlated yet very complementary information available in the various modalities present in narrated videos \cite{asano2020labelling}. 
To learn better feature extractors for audio and video, recent works, XDC \cite{alwassel2020self} and SeLaVi \cite{asano2020labelling} extend this clustering idea to the multimodal space. While these approaches focus on learning better feature extractors for each domain separately, our goal is to learn a joint multimodal embedding. As shown in Figure~\ref{fig:compare}, these cross-domain clustering methods (left) create separate clusters and use cross-domain pseudo-labels as the supervision for each feature extractor. In contrast, our model (right) creates a common embedding space across all modalities and performs clustering jointly.


 \begin{figure*}[tph]
 \centering 
  \includegraphics[width=2\columnwidth]{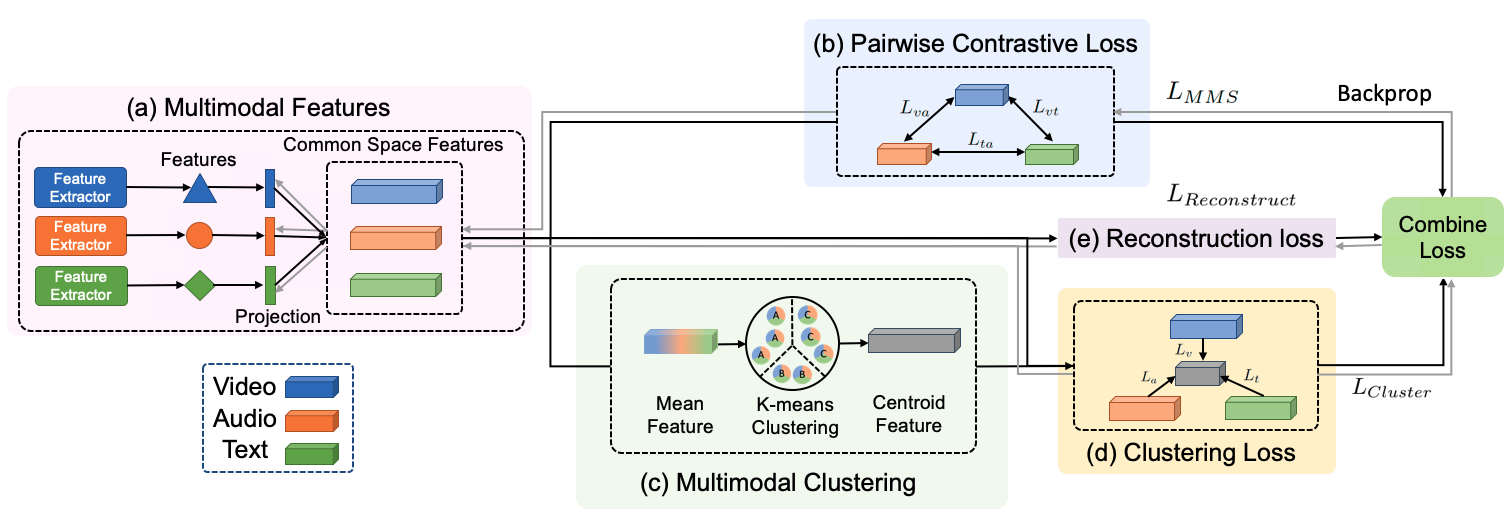}
\caption{\textbf{Illustration of our proposed framework.} Our framework comprises four parts: (a) Extracting features from several modalities and projecting them into joint space. (b) Calculating contrastive loss pairwise to pull the features close across modalities. (c) Performing multimodal clustering across features from different domains in a batch. (d) Performing joint prediction across features to multimodal centroids to bring together semantically similar embeddings. (e) Reconstruction loss for regularization. Best viewed in color.}
\label{fig:flow}
\vspace{-3mm}
\end{figure*}

\section{Learning to Cluster Multimodal Data}

To effectively construct a {\em joint representation space} from unlabeled narrated videos, we start with $n$ narrated video clips. Each video clip is associated with its corresponding visual representation, audio representation and text narration.
Given this input, the joint embedding space is learned, where the embeddings of video clips with semantically similar visual, audio, and text content are close to each other and apart when the content is dissimilar, as illustrated in Figure~\ref{fig:open}.

Using the notation in \cite{miech2020end}, for each clip, let video $v\in\mathcal{V}$ denote its visual representation,  $a\in\mathcal{A}$ represent its corresponding audio and $t\in\mathcal{T}$, its matching text narration generated using an automatic speech recognition (ASR) system.
Given a set of $n$ tuples of associated video, audio and text narrations $\{(v_i,a_i,t_i)\}_{i=1}^ n\in (\mathcal{V}\times\mathcal{A}\times\mathcal{T})^n$, as shown in Figure~\ref{fig:flow} (a), we first construct three parametrized mappings that derive embedding representations from the original video, audio and text signals. 
Transform $f:\mathcal{V}\to\mathbb{R}^d$ derives a $d$-dimensional embedding representation $f(v)\in\mathbb{R}^d$ from a video clip $v$, transforms $g:\mathcal{A}\to\mathbb{R}^d$ and $h:\mathcal{T}\to\mathbb{R}^d$, produce similar $d$-dimensional audio and text embeddings: $g(a)=z\in\mathbb{R}^d$ and $h(t)\in\mathbb{R}^d$.
In this work, $f$ takes as input pre-extracted 2D and 3D features from a fixed-length clip, the input for $g$ are log-mel spectrograms extracted from the audio segments, and for $h$, we use a sentence based neural model that transforms a set of words into a single vector.
More details about model architectures are in Section~\ref{sec:exp}.

Next, we introduce three loss functions to guide and properly situate these embeddings in the joint embedding space. A contrastive loss $L_{MMS}$ is used to ensure that the representations from each of the three modalities are comparable. A second clustering loss $L_{Cluster}$ encourages representations from semantically similar samples across all modalities to remain close in the learned embedding space. A third reconstruction loss $L_{Reconstruct}$ regularizes the multimodal common space features for more stable clustering training. The final model is trained to minimize  sum of these losses. 
\begin{align}
    \label{eq:total}
    \resizebox{.7\linewidth}{!}{$
    L = L_{MMS}+L_{Cluster}+L_{Reconstruct}
    $}
\end{align}

\subsection{Contrastive Loss for Learning Joint Spaces}
\label{sec:contrastive}
To learn a joint space for the three modalities, we compute a contrastive loss on all pairs of modalities, $(v,t),(t,a),(a,v)$, as shown in Figure~\ref{fig:flow} (b). 
This loss maximizes the similarity between representations corresponding to any two modalities from the same instance (video clip) while minimizing the similarity of imposter pairs from the two modalities from one clip of video to another. 
In this work, we use the Masked Margin Softmax (MMS)  function~\cite{ilharco2019large}, which defines the similarity between representations from two modalities in terms of their learned embedding vectors' dot product within a batch $B$.
Features from each of the three modalities $\{V,A,T\}$ are assembled for each batch. The total contrastive loss $L_{MMS}$ is the sum of pairwise losses using each of the three modalities:
\begin{equation}
    \label{eq:mms}
    \resizebox{.45\linewidth}{!}{$
    L_{MMS} = L_{ta}  + L_{vt} + L_{va} 
    $}
\end{equation}
where $L_{ta}$, $L_{vt}$, $L_{va}$ represent the loss associated with pairwise modalities $(t,a),(v,t),(a,v)$ respectively. For a pair of modalities, for example the text and audio modalities, the individual loss $L_{ta}$ is in turn given as:
\begin{align}
    \resizebox{0.9\linewidth}{!}{$
    L_{ta} = -\frac{1}{B} \sum\limits_{i=1}^{B} \Bigg[ \Bigg( \log \frac{\displaystyle e^{h(\textbf{t}_i) \cdot g(\textbf{a}_i) - \delta}}{\displaystyle e^{h(\textbf{t}_i) \cdot g(\textbf{a}_i)  - \delta} + \textstyle \sum\limits_{\substack{k=1 \\ k\neq i}}^{B}e^{h(\textbf{t}_k^{imp}) \cdot g(\textbf{a}_i)}} \Bigg)
    $} \\       \nonumber 
    \resizebox{0.7\linewidth}{!}{$
    + \Bigg( \log \frac{\displaystyle e^{h(\textbf{t}_i) \cdot g(\textbf{a}_i) - \delta}}{\displaystyle e^{h(\textbf{t}_i) \cdot g(\textbf{a}_i)  - \delta} + \textstyle \sum\limits_{\substack{j=1 \\ j\neq i}}^B{e^{h(\textbf{t}_i) \cdot g(\textbf{a}_j^{imp})}}} \Bigg)\Bigg]
    $} \nonumber
\end{align}
where $a_j^{imp}$ represents imposter pairs from two modalities that are sampled from a batch but do not co-occur. As can been seen in the $L_{ta}$ case, this loss attempts to discriminate between positive or true embedding pairs and imposter or negative pairs within each batch. Using two separate parts, the space of positive and negative samples is enumerated separately: in one case, a given text sample is paired with various negative audio samples. In the second case, an audio sample is paired with various negative text samples. ($i$, $j$, $k$) are various indices of video clips in a given batch.  $\delta$ is a margin hyperparameter that is empirically selected. 
By projecting all features to the same space and ensuring that their similarities are maximized pairwise, this formulation of the pairwise contrastive loss ensures that the features across different modalities are comparable. 

\subsection{Clustering Multimodal Features}

To ensure that representations of semantically related instances are close in the learned joint multimodal space, in addition to contrastive loss described above, a self-supervised clustering step is included as part of the training process. 

\noindent \textbf{Online K-means clustering.}
We applied standard clustering algorithm $k$-means 
that takes a set of vectors as input, in our case, the features $M$ produced by the fused multimodal feature:
\begin{equation}
    \label{eq:fused}
    \vspace{-1mm}
    \resizebox{.5\linewidth}{!}{$
    M =  (f(\textbf{v}) + g(\textbf{a}) + h(\textbf{t})) / 3
    $}
\end{equation}
where we take the mean over embeddings from three modalities to represent a multimodal instance. We cluster them into $k$ distinct groups.
More precisely, it outputs a $d\times k$ centroid matrix $C=\{\mu_1,..,\mu_k\}$ and the cluster assignments $y_n$ of each multimodal instance $n$ are defined by solving the following problem:
\begin{equation}
\label{eq:kmeans}
  \min_{C \in \mathbb{R}^{d\times k}}
  \frac{1}{N}
  \sum_{n=1}^N
  \min_{y_n \in \{0,1\}^{k}}
  \| M_n -  C y_n \|_2^2
\end{equation}
We then acquire a centroid matrix $C^*$ and a set of assignments $(y_n^*)_{n\le N}$.
Unlike pseudo-labels-based methods \cite{caron2018deep} that only make use of the assignments (labels), we make use of the centroid matrix for semantic learning.
To cover variant semantic information for clustering, we use features from the previous batches to gather sufficient instances for online learning.

\noindent \textbf{Semantic centroid learning.}
To learn the features closer to its multimodal semantic centroids. We proposed to use the centroid as a contrastive loss reference target. This target pulls the features from three modalities closer to the centroid that is close to their multimodal instance feature $M_n$ and pushes the features far away from the other centroid. For each modality, for example, the text modalities, the individual loss $L_t$ is in turn given as:
\vspace{-1mm}
\begin{align}
    \resizebox{0.5\linewidth}{!}{$
    L_{t} = -\frac{1}{B} \sum\limits_{i=1}^{B}  \log \frac{\displaystyle e^{h(\textbf{t}_i) \cdot \mu'  - \delta}}{\displaystyle  \textstyle \sum\limits_{\substack{k=1 }}^{K}e^{h(\textbf{t}_i)\cdot \mu_k} } 
    $}        
\end{align}
where $\mu'$ is the nearest centroid for the multimodal instance feature $M_i$ and $\mu'$. We later sum over the loss from three modalities:
\vspace{-1mm}
\begin{equation}
    \label{eq:cluster}
    \resizebox{.45\linewidth}{!}{$
    L_{Cluster} = L_{v}  + L_{a} + L_{t} 
    $}
\end{equation}

In the end, the projected features learn to be closer to its centroid feature among the three and also learns to be closer in similar semantics.

\noindent\textbf{Multimodal features reconstruction.}
Reconstruction can help in capturing features that are suppressed by contrastive learning/clustering \cite{chen2020intriguing}. In a video of \textit{chopping onions}, with both the sound of chopping in the background as well as the speech/text with the word \textit{onion} in the foreground, it is possible that contrastive learning/clustering will focus more on associating the video with either the sound (background) or the speech (foreground), but not both. We hypothesize that the reconstruction loss will force the capture of features from both background and foreground, which is important for retrieval/other downstream tasks. Reconstruction is also an auxiliary task that helps regularize training and improve generalization~\cite{le2018supervised}.
We performed a reconstruction loss on top of the common space features from three modalities to stabilize the feature training during clustering. 
For each modality, for example, the visual modalities, the individual loss $L_{v'}$ is in turn given as:
\begin{equation}
    \label{eq:reconstruct_single}
    \resizebox{.55\linewidth}{!}{$
    L_{v'} = -\frac{1}{B} \sum\limits_{i=1}^{B} \norm{ f'(\textbf{v}) -f(\textbf{v}) } ^2 $}
\end{equation}
where $f'(\textbf{v})$ represented the reconstructed features by feeding $\textbf{v}$ into two linear layers as encoder and decoder.
We then sum the loss over each modality:
\begin{equation}
    \label{eq:reconstruct}
    \resizebox{.55\linewidth}{!}{$
    L_{Reconstruct} = L_{v'}  + L_{a'} + L_{t'} 
    $}
\end{equation}

\section{Experiments}
\label{sec:exp}
\subsection{Implementation details}

For the visual branch of the proposed MCN model we follow \cite{miech2019howto100m} and use pre-trained 2D features from a ResNet-152 model~\cite{he2016deep} trained on ImageNet \cite{deng2009imagenet} to extract features at the rate of one frame per second, along with pre-trained 3D features from a ResNeXt-101 model~\cite{hara2018can} trained on Kinetics~\cite{carreira2017quo} to obtain 1.5 features per second. 
The video clip features were computed by concatenating the 2D and 3D features into a 4096 dimension vector and max-pooling the features over time.
For the audio branch of the network, we compute log-mel spectrograms and use a pre-trained DAVEnet model \cite{harwath2018jointly} to extract audio features. For the textual branch, the feature extraction process proposed in ~\cite{miech2019howto100m} is adopted to extract text representations: a GoogleNews pre-trained Word2vec model~\cite{mikolov2013efficient} provides word embeddings, followed by a max-pooling over words in a given sentence to extract a sentence embedding.  Note that all backbones are fixed, and they are not fine-tuned during training. Each feature extraction branch is followed by a separate fully-connected layer and a gated unit for projecting the features in a common embedding space. To allow for pairwise comparisons, features from each of the different modalities are set to be 4096-dimensional vectors. More details can be found in the supplement.


\subsection{Datasets}
\noindent \textbf{Training Dataset.}
Our models are trained on the HowTo100M~\cite{miech2019howto100m} instructional video dataset, which contains 1.2M videos along with their corresponding audio that consists of speech and environmental sound and automatically generated speech transcriptions.

\noindent \textbf{Downstream Datasets.}
The {\bf YouCook2}~\cite{zhou18towards} dataset contains 3.5K cooking instruction video clips with text descriptions collected from YouTube. Unlike Howto100m dataset, text descriptions in YouCook2 are human-annotated. 
The {\bf  MSR-VTT}~\cite{xu16msrvtt} dataset contains 200K human annotated video clip-caption pairs on various topics. We use the same test set with 1K video clip-caption pairs constructed in~\cite{miech2019howto100m} in our experiments.
\noindent The {\bf CrossTask}~\cite{zhukov2019crosstask} dataset contains 2.7K instructional videos that cover various topics.
The action steps and their order for each task were collected from \textit{wikiHow} articles with manual annotation for each frame. 
\noindent The {\bf  Mining Youtube}~\cite{kuehne2019mining} dataset
focuses on YouTube videos for five simple dishes.
The test set contains 250 cooking videos, 50 of each task, that are densely annotated, \ie each frame is labeled with its respective action class.

\begin{table}
		\tablestyle{2pt}{1.05}
		\resizebox{\columnwidth}{!}{
		\begin{tabular}{@{}l|c|cc|ccc|ccc@{}}
			\toprule
			 \multicolumn{4}{c}{} & \multicolumn{3}{c}{YouCook2}                  & \multicolumn{3}{c}{MSRVTT}                   \\ 
			\cmidrule(lr){5-7} \cmidrule(lr){8-10} 
			Method        & Mod & Model  &  TR & R@1  & R@5  & R@10 & R@1 & R@5  & R@10 \\ \midrule
			Random & & - & - & 0.03 & 0.15 & 0.3 & 0.01& 0.05& 0.1 \\
			Miech  \cite{miech2019howto100m} & VT &  R152+RX101  & N  & 6.1 & 17.3 & 24.8                        & 7.2 & 19.2 & 28.0      \\
            MDR \cite{amrani2020noise} & VT & R152+RX101 & N & - & - &- & 8.0 & 21.3 & 29.3 \\
            MIL-NCE* \cite{miech2020end}& VT  & R152+RX101 & N  & 8.1 & 23.3 & 32.3                        & 8.4 & 23.2 & 32.4                      \\
            \textbf{MCN (ours)} & VAT & R152+RX101 & N & \textbf{18.1} & 35.5 & 45.2                        & \textbf{10.5} & \textbf{25.2} & \textbf{33.8}                        \\
            \midrule
            MDR \cite{amrani2020noise} &VT & R152 & N & - & - &- & 8.4 & 22.0 & 30.4 \\
			ActBERT \cite{zhu2020actbert} &VT & R101+Res3D & N  & 9.6 & 26.7 & 38.0                        & 8.6 & 23.4 & 33.1                      \\
			SSB \cite{patrick2020support}& VT  & R(2+1)D-34+R152 & N  & - & - & -                        & 8.7 & 23.0 & 31.1                      \\
            
 \midrule
            MMV FAC  \cite{alayrac2020self}          &VAT & TSM-50x2 & Y   & 11.7 & 33.4 & 45.4             & 9.3 & 23.0 & 31.1                      \\
			MIL-NCE~\cite{miech2020end} &VT & I3D-G      & Y & 11.4 & 30.6 & 42.0  & 9.4 & 22.0 & 30.0                    \\
			MIL-NCE~\cite{miech2020end} &VT & S3D-G      & Y & 15.1 & \textbf{38.0} & \textbf{51.2}  & 9.9 & 24.0 & 32.4                    \\
			
			\bottomrule
			 
		\end{tabular}
		}
		\caption{Comparison of text-to-video retrieval systems. Mod indicates modality used, where V: video, A: audio, T: text. TR indicates if a trainable backbone is used or not. 
		\label{tab:retrieval}
		}
		\vspace{-0.25cm}
\end{table}

\begin{table}
		\tablestyle{2pt}{1.05}
		\resizebox{\columnwidth}{!}{
		\begin{tabular}{@{}l|c|cc|ccc|ccc@{}}
			\toprule
			 \multicolumn{4}{c}{}& \multicolumn{3}{c}{CrossTask} & \multicolumn{3}{c}{MYT}   \\ 
			\cmidrule(lr){5-7} \cmidrule(lr){8-10} 
			Method  & Mod   & Model & TR   & Recall & IOD & IOU & Recall & IOD & IOU \\ 
			\midrule
			CrossTask \cite{zhukov2019crosstask}  &VT & R152+I3D & N  & 22.4 & - & -  & -  & - & -  \\
			CrossTask \cite{zhukov2019crosstask}  &VT & R152+I3D & N  & 31.6  & - & - & -  & - & -  \\
			Mining: GRU \cite{kuehne2019mining}  &VT  & TSN & N  & -  & - & -  &   - & 14.5 & 7.8  \\
			Mining: MLP \cite{kuehne2019mining}  &VT & TSN & N   & -  & - & -   & - & 19.2 & 9.8  \\
			\midrule
			Miech  \cite{miech2019howto100m}  &VT  & R152+RX101 & N   & 33.6 & 26.6 & 17.5   & 15.0 & 17.2 & 11.4 \\
			MIL-NCE* \cite{miech2020end} &VT & R152+RX101 & N & 33.2 & 30.2 & 16.3   & 14.9 & 26.4 & 17.8                      \\
            
			\textbf{MCN (ours)}  &VAT & R152+RX101 & N & 35.1 & \textbf{33.6} & \textbf{22.2}                        & \textbf{18.1}& \textbf{32.0} & \textbf{23.1}                        \\
			\midrule
			ActBERT \cite{zhu2020actbert}  &VT & R101+Res3D & N & 37.1 & - & -   &  - & - & - 
			\\
			ActBERT \cite{zhu2020actbert}  &VT & + Faster R-CNN & N & \textbf{41.4} & - & -   &  - & - & - 
			\\
			\midrule

			  MIL-NCE \cite{miech2020end}  &VT & I3D-G & Y & 36.4 & - & -   & - & - & -                      \\
			  MIL-NCE \cite{miech2020end}  &VT & S3D-G & Y & 40.5 & - & -   & - & - & -                      \\
			
			\bottomrule
		\end{tabular}
		}
		\caption{Evaluation of temporal action localization systems.  
		\label{tab:temporal_state}
		}
		\vspace{-0.45cm}
\end{table} 

\subsection{Downstream Tasks}
To demonstrate the effectiveness of the proposed model, we evaluate embeddings derived from the network in two downstream tasks: text-to-video retrieval and temporal action localization.
We focus on the zero-shot task because we want to access the quality of the cross-modal semantic embedding that was learned during training.
When performing retrieval using our model, we compare the query text features with the video and audio features by computing similarity for both and using the average. For action localization, we compute the same distance of the video-audio pair of each frame to each respective label embedding and are so able to align video frames to each of the provided action steps. 

\noindent\textbf{Text-to-Video Retrieval.}
The goal of this task is to retrieve the matching video from a pool of videos, given its ground truth text query description.  
The model is tested on two video description datasets and evaluated on recall metrics: R@1, R@5, R@10. These evaluations are used to demonstrate the effectiveness of the contrastive loss and learned joint embedding space across three modalities. 

\noindent\textbf{Text-to-Full Video Retrieval.} The conventional text-to-video retrieval task attempts to match a caption (or ground-truth text query) to a single video clip. Since a single caption can refer to many individual clips within a dataset, this task is limiting. To this end, we propose the task of \textit{text-to-full video retrieval} where the goal is to match a set of captions (or text queries) describing multiple parts of a video to an entire video. This is a more realistic task than single clip retrieval since various real-world applications require retrieving entire videos from complex textual queries. We evaluate on YouCook2 dataset with recall metrics: R@1, R@5, R@10.

\noindent\textbf{Temporal action localization.}
We further evaluate our model on two temporal action localization tasks. 
The CrossTask \cite{zhukov2019crosstask} dataset considers the task of clip level action detection. Here, an unordered set of action labels is given for a set of clips of the same video, and clips have to be classified with the respective action labels. The performance is reported as recall and computed as a ratio of the correctly predicted clips over the total number of clips in the video as used in \cite{zhukov2019crosstask}. 
The MiningYoutube \cite{kuehne2019mining} dataset considers the task of frame-level temporal action segmentation.
Here, each test video is provided together with the respective actions and their ordering, including the background. The goal is to find the correct frame-wise segmentation of the video given the action order. We follow the inference procedure outlined in \cite{kuehne2019mining} to compute the alignment given our similarity input matrix. The dataset employs two evaluation metrics: intersection over detection (IoD) \cite{bojanowski2014weakly}, defined as $\frac{G \cap D}{D}$: the ratio between the intersection of ground-truth action $G$ and prediction $D$ to prediction $D$, and the Jaccard index, which is an intersection over union (IoU) given as $\frac{G \cap D}{G \cup D}$. 


\subsection{Comparison with State-of-the-art Methods}

\noindent
\textbf{Zero-shot Video Retrieval. }
We first examine the results of the text-to-video retrieval task on the YouCook2 and MSR-VTT datasets (Table~\ref{tab:retrieval}).
We compare only with baseline models that were not fine-tuned on the respective dataset for a fair comparison. 
To allow comparability between different approaches, we use a fixed visual feature extraction backbone as described in \cite{miech2019howto100m} whenever possible. For the baseline MIL-NCE* \cite{miech2020end}, we apply their training strategy on the same visual feature set we use, ResNet-152 (R152) and ResNeXt-101 (RX101) \cite{miech2019howto100m}.
On YouCook2, our model significantly outperforms prior works on the same architecture and shows even competitive results compared to models with trainable visual backbone (TR). 
Our method also performs better than the other baselines on MSR-VTT. The gains are, however, not as significant as on YouCook2. We attribute this to the fact that neither the available audio nor the textual description is instructional in nature and, therefore, semantically further away from our training set. 

\noindent\textbf{Zero-shot Action Localization. }
We examine the action localization tasks on the CrossTask and the MiningYouTube dataset in Table~\ref{tab:temporal_state}.
For CrossTask, given each frame in the video, we perform a zero-shot classification of the given labels and calculate the recall. In this zero-shot setting, the model computes video text similarity to localize action step labels similar to~\cite{miech2019howto100m}.
Our method outperforms state-of-the-art approaches for self-supervised learning \cite{miech2020end,miech2019howto100m} and a fully supervised approach \cite{zhukov2019crosstask} especially in the IOU and IOD metrics, which also consider false-positive predictions from the background class as an action step. 
Approaches in \cite{miech2019howto100m} and MIL-NCE* \cite{miech2020end} are directly comparable with our method since they use the same feature extractor as us. In contrast, MIL-NCE \cite{miech2020end} uses a stronger video backbone and \cite{zhu2020actbert} uses additional feature modalities such as region features along with a stronger language model.
We also evaluate our model on the MiningYoutube~\cite{zhukov2019crosstask} temporal action localization benchmark. 
Our method outperforms state-of-the-art approaches for both self-supervised \cite{miech2020end, miech2019howto100m} and weakly supervised  \cite{kuehne2019mining} learning. More settings, including data and computing resources for each model, are in the supplement.

\noindent \textbf{Clustering Metrics. }
We further evaluate our system with respect to various clustering metrics as proposed by \cite{asano2020labelling}. Results are shown in Table \ref{eq:cluster_result}. The definition of each metric is included in the supplement.  
It shows that our learned multimodal features are closer to the ground-truth distribution and have higher purity within the cluster.

\begin{table}[t]
    \centering
    \tablestyle{2pt}{1.05}
  \begin{tabular}{l @{} c c c c c}
  \toprule
   \multicolumn{2}{c}{}& \multicolumn{3}{c}{CrossTask} & \multicolumn{1}{c}{}   \\ 
  \midrule
     Method                        &   \textbf{NMI} $\uparrow$ & \textbf{ARI} $\uparrow$ & \textbf{Acc.} $\uparrow$& $\langle \mathbf{H} \rangle$ $\downarrow$ & $\langle \mathbf{p_\mathrm{max}}  \rangle$ $\uparrow$ \\
    \midrule 
    Random     &  3.2 & 3.2 & 9.4 & 1.30 & 47.5 \\
    Miech \etal \cite{miech2019howto100m}        & 61.8 & 46.1 & 57.0 & 0.39 & 81.5 \\
    MIL-NCE* \cite{miech2020end}   & 62.0 & 45.6 & 56.7 & 0.37 & 82.4\\
    \midrule
    \textbf{MCN (ours)} & \textbf{65.5} & \bf{48.5} & \textbf{57.6} & \textbf{0.34} & \textbf{83.8}  \\
    \bottomrule
  \end{tabular}
  \caption{Performance on clustering metrics on the CrossTask dataset evaluated by GT text annotations on video segments. \label{eq:cluster_result}}
  \vspace{-0.65cm}
 \end{table}


 \begin{figure*}[tbhp]
 \centering
  \includegraphics[width=1.75\columnwidth]{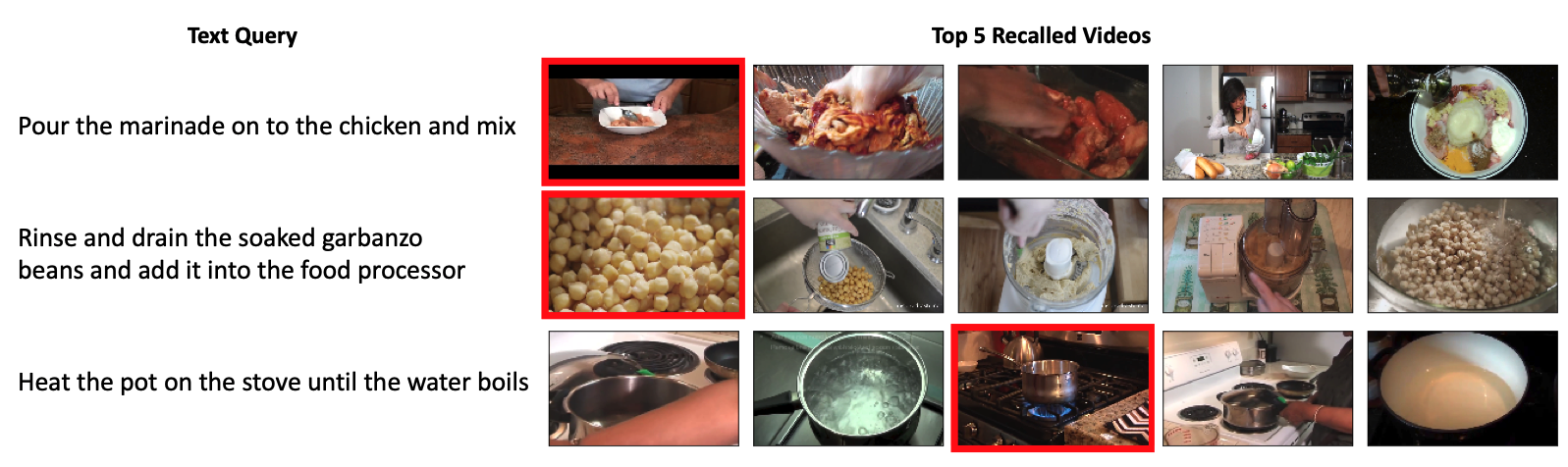}
  \vspace{-0.25cm}
\caption{Qualitative results for the text-to-video retrieval task on YouCook2. Top-ranked clips show a high similarity to the described task as well as among each other without being too visually similar. \label{fig:retrieve_visual}}
\vspace{-0.35cm}
\label{fig:retrieval}
\end{figure*}

\begin{figure}[tbhp]
 \centering
  \includegraphics[width=\columnwidth]{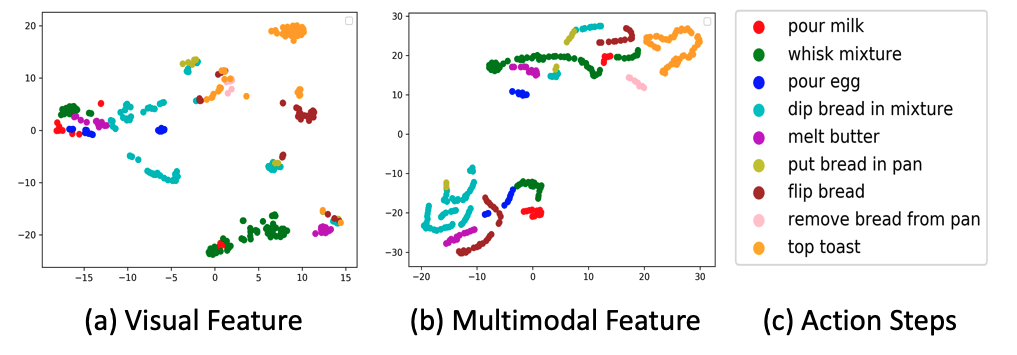}
  \vspace{-5mm}
\caption{t-SNE visualizations on the CrossTask dataset for the task of "Make French Toast". Best viewed in color.}
\vspace{-0.15cm}
\label{fig:cluster_visual}
\end{figure}

\subsection{Full Video Retrieval}

To address the problem of full video retrieval from a set of captions, we divide each video into a set of clips, which are compared with the queries. We evaluate three different methods: 
In \textbf{majority vote over clip} predictions, we obtain the top-k predictions of each clip/caption pair as votes and select the video which has the majority of votes. For \textbf{majority vote over videos}, the maximal prediction over all the clips of a video is taken for each caption to obtain video/caption pairs. Then, the top-k of these predictions are selected as votes, and the video with the most votes is predicted. Lastly, our \textbf{caption averaging} method involves obtaining the maximal prediction over all the clips of a video is taken for each caption and then averaging over the set of captions in a query. This gives a single prediction for the entire video. 

We examine the results of the text-to-full video retrieval task on the YouCook2 dataset (Table \ref{tab:fullvideoretrieval}). Of the three methods to obtain full video predictions, the caption averaging achieves better results than both majority voting schemes. Furthermore, we find that our method outperforms prior works on this task with a 6.8\% improvement on R@1. Since we obtain full video predictions, we also perform full-video classification on the CrossTask dataset using the set of sub-task labels as the set of query captions, where we achieve a top-1 accuracy of 68.7\%.

\begin{table}[t]
        \centering
        \small
        \begin{tabular}[t]{@{}lcccc@{}}
			\toprule
				Method   & Prediction & R@1 & R@5 & R@10 \\
				\midrule
				Random  &  - &  0.23 & 1.15 & 2.32  \\
				\midrule
				MCN (ours) & MV-Clip     &  38.8 & 67.4 & 76.8  \\
				MCN (ours) & MV-Video    &  38.8 & 67.7 & 78.4  \\
				MCN (ours) & Caption Avg.  & \textbf{53.4} & \textbf{75.0} & 81.4 \\
				\midrule
    Miech \etal \cite{miech2019howto100m}  & Caption Avg.      & 43.1 & 68.6 & 79.1  \\
    MIL-NCE* \cite{miech2020end} & Caption Avg.  & 46.6 & 74.3 & \textbf{83.7} \\
				 
			\bottomrule
		\end{tabular}
		
		\caption{Comparison of Text-to-Full Video retrieval systems on the YouCook2 dataset. The prediction column denotes the method used to obtain video-level predictions: majority vote over clips (MV-Clip), majority vote over videos (MV-Video), and caption averaging (Caption Avg.). \label{tab:fullvideoretrieval}}
\end{table}

\subsection{Ablation Studies}
                  
\begin{table}[t]
        \centering
        \resizebox{\columnwidth}{!}{
		 \begin{tabular}[t]{@{}lcccc@{}}
			\toprule
				Loss   & \rYC & \rMSRVTT & \rCrossTask & MYT-IOU \\
				\hline
				NCE                           & 39.2   & 33.5  & 33.9 &   21.5 \\
				MIL-NCE                       & 40.0  & 33.0  & 33.7 &   21.1 \\
				MMS                           & 43.7  & 32.9  & 34.3 &   22.1 \\ 
				MMS + Cluster                 & 44.3 & 33.7  & 34.5 &   22.6 \\ 
				\midrule
				MMS + Cluster + Reconstruct   & 45.2  & 33.8  & 35.1 &   23.1 \\ 
			\bottomrule
		\end{tabular}
		}
		
		\caption{Ablation study on different loss including the selection of contrastive learning loss, the additional clustering, and reconstruction loss. \label{tab:loss_abla}}
		\vspace{-0.35cm}
\end{table}

\begin{table}[t]
        \centering
        \resizebox{\columnwidth}{!}{
		 \begin{tabular}[t]{@{}lcccccc@{}}
			\toprule
				Method  & Target & Labels & \rYC & \rMSRVTT & \rCrossTask & MYT-IOU \\
				\hline
				Sinkhorn  & Swap  & hard & 39.0  & 33.4  & 33.6 &   21.1 \\ 
				Sinkhorn  & Swap  & soft & 41.8  & 33.9  & 34.5 &   22.1 \\
				Sinkhorn & Joint  & hard & 44.4  & 33.4  & 34.6 &   21.1 \\ 
				Sinkhorn & Joint  & soft & 43.6 & 32.4  & 34.1 &   21.6 \\ 
				
				K-means & Swap  & hard & 41.3  & 32.8  & 33.2 &   21.0 \\ 
				K-means & Joint  & hard & 44.3 &   33.1  & 34.6 &   21.4 \\
				\midrule
				K-means & Centroid & hard & 45.2 &   33.8  & 35.1 &   23.1 \\
			\bottomrule
		\end{tabular}
		}
		
		\caption{Ablation study on different clustering pipelines with various methods, loss prediction target, and label types. \label{tab:cluster_compare}}
		\vspace{-0.45cm}
\end{table}









\label{subsec:ablation}
To better understand the contributions of various algorithmic design choices used to build the proposed MCN model, we perform a set of ablation studies on the following downstream tasks: YouCook2 R@10 (\rYC), MSR-VTT R@10 (\rMSRVTT),  CrossTask average recall (\rCrossTask) and MiningYoutube IOU (MY-IOU). For each setting, we use the same feature extractor for three modalities as described in Sec 4.1 for a fair comparison. More ablations are in the supplement.

\vspace{2mm}

\noindent\textbf{Selection on different losses.}
In our first set of experiments, we find the proposed clustering is crucial not only for clustering-related tasks but also for retrieval (MSR-VTT) tasks as shown in Table \ref{tab:loss_abla}. This validates our hypothesis that semantically close instances should be clustered closely in the joint embedding space. Also, the selection of contrastive loss (MMS) shows better results in our model.

\noindent\textbf{Different choices of clustering methods.}
We evaluate the performance of (1) Selection of different clustering methods such as Sinkhorn clustering \cite{asano2019self} and K-means \cite{arthur2006k}. (2) Different prediction targets such as using swap prediction, which uses the pseudo label of other modalities for prediction target as \cite{caron2020unsupervised,alwassel2020self}. Or using the mean feature pseudo label as a joint prediction for three modalities. Also, using the centroid of the cluster as the target. (3) Different prediction labels, including hard labels (one-hot) or soft labels (continuous). Detailed descriptions are included in the supplement.  As shown in Table \ref{tab:cluster_compare},
our method encourages each modality feature to move closer to the semantic centroid, which improves performance by explicitly encouraging semantically close features from different domains to cluster together. 


\subsection{Qualitative Analysis}
We perform a qualitative analysis with the model's ability to do zero-shot text-to-video retrieval shown in Figure \ref{fig:retrieve_visual}. Given an open-vocabulary caption, our model can retrieve the correct corresponding video segment. 
We also visualize the efficacy of using multimodal embeddings (concatenated video and audio representations) over using only visual embeddings. Representations from the CrossTask dataset are visualized using t-SNE plots. We observe that with multimodal features as Figure~\ref{fig:cluster_visual} (b), semantically related instances (based on ground truth classes) tend to be more tightly related than uni-modal visual features trained from contrastive loss (a) that appear more spread out. Also, multimodal features are clearly more separable for different actions. 

\section{Conclusions}

We have developed a novel self-supervised multimodal clustering network that learns a common embedding space by processing local (via a contrastive loss) and global (via a clustering loss) semantic relationships present in multimodal data. The multimodal clustering network is trained on a large corpus of narrated videos without any manual annotations. 
Our extensive experiments on multiple datasets show that creating a joint video-audio-language embedding space with a clustering loss is essential for self-supervised learning of good video representations.
Our approach can be extended to more modalities such as optical flow or sentiment features and applied to other multimodal datasets for learning joint representation spaces without human annotation.

{\scriptsize \noindent\textbf{Acknowledgments}: We thank IBM for the donation to MIT of the Satori GPU cluster.
This work is supported by IARPA via DOI/IBC contract number D17PC00341. The U.S. Government is authorized to reproduce and distribute reprints for
Governmental purposes notwithstanding any copyright annotation thereon.
 Disclaimer: The views and conclusions contained herein are those of the authors and should not be interpreted as necessarily representing the official policies or endorsements, either expressed or implied, of IARPA, DOI/IBC, or the U.S. Government.}


\clearpage
\appendix
\begin{center}
  {\large \bf Appendix \par}
\end{center}

\numberwithin{equation}{section}
\setcounter{table}{0}
\setcounter{figure}{0}
\setcounter{equation}{0}

\noindent This appendix is organized as follows: \\
A. Method description and comparisons. \\
B. Details on experimental setups. \\
C. Additional experiment results. \\
D. Quantitative experimental results for text-to-video retrieval and temporal action localization.

\section{Method description and comparisons.}

\subsection{Salient features of the MCN method}

To highlight various aspects of our proposed MCN method, we compare our method with another notable multimodal cluster method: XDC \cite{alwassel2020self}.\newline
\textbf{Goal of the model}. While XDC's goal is to learn representations for each modality, with the MCN method, we try to learn a joint representation across modalities. The two approaches are hence complementary, given that they target different tasks. In addition, XDC aims to learn feature backbones from scratch since these feature backbones will be applied to single modality downstream tasks. In contrast, we start from pre-trained feature extractors and aim to learn projection heads across domains to derive a joint space from the three modalities.\newline 
\textbf{Joint space of representation. } Based on the formulation of XDC, pseudo-labels from one modality serve as prediction targets of another. Since the prediction target for the visual and audio instances are different, the model will not learn a joint space across modalities. The paper also proposed a CDC method where the prediction target of visual and audio instances are the same. However, it is not evaluated on multimodal tasks. \newline
\textbf{Combining contrastive learning. }
While XDC uses only the clustering loss, we combine multiple losses together. We find the contrastive loss to be crucial in multimodal tasks since it pulls the instances across modalities that co-occur together. In general, this supervision is crucial in most multimodal pre-training strategies. \newline
\textbf{Use of different modalities}
Since our goal is to learn a joint space across three modalities, our motivation for using audio is slightly different from XDC. XDC uses audio and video as complimentary learning signals for self-supervised prediction targets. On the other hand, we find audio as a modality that bridges the gap between video and text, since audio and video preserve fine-grained information. The text modality represents a more abstract concept, distilled from the audio signal using ASR. Hence, we find learning from the three modalities to be beneficial.


\section{Experiment Details}

\begin{figure*}[tbhp]
\centering
\includegraphics[width=1.95\columnwidth]{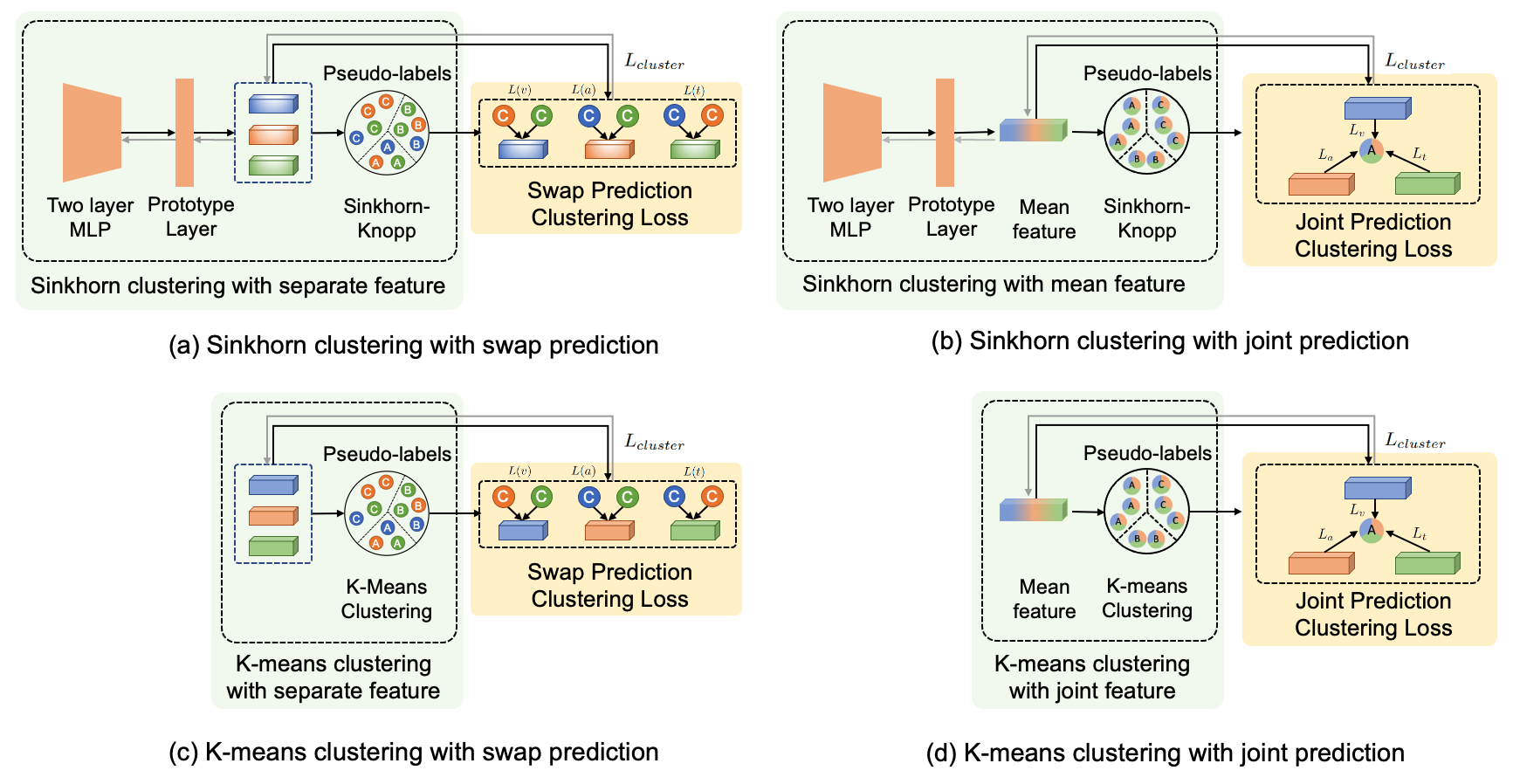}
\caption{Comparison of different clustering pipelines. We investigate different clustering pipelines in replace of the clustering loss in our main paper. (a) Performs a sinkhorn clustering folloing a swap prediction. The loss was calculated between the clustered features and pseudo labels. (b) Replaces the swap prediction to joint prediction by performing the clustering on the mean feature. The loss was calculated by the mean pseudo label and the projected feature in Figure 3a. (c) Performs K-means along with swap prediction. (d) Performs K-means on the mean features and performs joint prediction. }
\label{fig:compare_cluster2}
\end{figure*}
\subsection{Implementation details}

We use an Adam optimizer~\cite{kingma2015adam} with a learning rate of $1\mathrm{e}{-4}$ and cosine learning rate schedule \cite{misra2020self}. The model is trained for 30 epochs on four V100 GPUs over a period of about two days. Various hyperparameters in our experiments are set as follows: margin hyperparameter $\delta = 0.001$ 
, and a batch size of $B$ = 4096 video clips and cluster size is set to be 256.

\subsection{Clustering metrics}

To better evaluate our learned features, we use the \textit{k}-means clustering algorithm and calculate various clustering metrics based on  ground-truth labels on the CrossTask \cite{zhukov2019crosstask} and MiningYouTube \cite{kuehne2019mining} tasks. In this case, the number of clusters \textit{k}, also corresponds to the number of possible steps assigned to the temporal action localization task for each video during test time. 

We follow the evaluation protocol and notations used in \cite{asano2020labelling} and report performance based on the following standard clustering metrics:  \emph{normalized mutual information} (NMI) \cite{strehl2002cluster},  \emph{adjusted rand index} (ARI) \cite{hubert1985comparing}, and \emph{accuracy} (Acc). These results are obtained  after matching the estimated \textit{k}-means pseudo-labels to the ground truth targets using the Kuhn–Munkres/Hungarian algorithm~\cite{kuhn1955hungarian}.
We also report the \emph{mean entropy per cluster}  :
\begin{equation}
    \langle H \rangle = \frac{1}{K}\sum_{k \in K} H(p(y \vert \hat{y}_k = k)), \label{eq:entropy}
\end{equation}
where $\hat{y}$ corresponds to the psuedo-labels generated by clustering and $y$ relates to the ground-truth labels. In this formulation $p(y \vert \hat{y}_k = k)$ denotes the distribution of ground-truth labels that fall in the generated clusters $k$, while $H(U)$ represents the entropy given as $- \sum_{i=1}^{|U|}P(i)\log(P(i))$.
In ideal conditions, the perfect mean entropy will be zero. 

We also report the the \emph{mean maximal purity per cluster},
\begin{equation}
    \langle p_\mathrm{max} \rangle = \frac{1}{K}\sum_{k \in K} \max(p(y \vert \hat{y}_k = k)), \label{eq:purity}
\end{equation}
In ideal conditions, the perfect mean purity will be $100\%$.

By using the various metrics described above, the clustering result on MiningYoucook dataset was shown in Table \ref{eq:cluster_result}. 
The overall results show a similar pattern with the experiment shown in the main paper using CrossTask dataset. 

\begin{table}[t!]
    \centering
    \tablestyle{4pt}{1.05}
  \begin{tabular}{l @{} c c c c c}
  \toprule
     Method                        &   \textbf{NMI} $\uparrow$ & \textbf{ARI} $\uparrow$ & \textbf{Acc.} $\uparrow$& $\langle \mathbf{H} \rangle$ $\downarrow$ & $\langle \mathbf{p_\mathrm{max}}  \rangle$ $\uparrow$ \\
    \midrule 
    Random     &  0.4 & 0.4 & 8.6 & 2.3 & 25.5 \\
    Miech \etal \cite{miech2019howto100m}        & 72.9 & 45.4 & 59.8 & 0.44 & 79.5 \\
    MIL-NCE* \cite{miech2020end}   & 73.1 & 46.8 & 60.6 & 0.37 & 77.9\\
    \midrule
    \textbf{MCN} & \textbf{75.8} & \bf{48.0} & \textbf{61.7} & \textbf{0.40} & \textbf{80.8}  \\
    \bottomrule
  \end{tabular}
  \caption{Performance on various clustering metrics for the MiningYouTube task \label{eq:cluster_result}}
 \end{table}

\subsection{Clustering ablation explanation}
In this ablation study, we investigate different clustering pipelines. Here, we break down our results into several categories and provide an analysis for various clustering methods, different prediction targets, and several kinds of pseudo-labels. 

\noindent \textbf{Clustering method. }
The goal of this analysis is to create various kinds of pseudo-labels as  prediction targets. If a pseudo-label can be thought of as a certain semantic representation of a cluster,  two instances that have the same pseudo-label, can then be considered as semantically similar.
The K-means method follows the deep clustering \cite{caron2018deep} approach which utilizes K-means clustering to create pseudo labels as prediction targets. These targets are then used for single modality learning on ImageNet \cite{russakovsky2015imagenet}.
The Sinkhorn clustering method follows the SeLa \cite{asano2019self} technique that utilized a trainable network to replace the K-means clustering for generating pseudo-labels. The method also applies an optimal transport  sinkhorn algorithm \cite{cuturi2013sinkhorn} to guarantee uniform distribution over different cluster labels, which in turn prevents the learnable clustering network (2 layers MLP) from learning a degenerated solution. More details of this sinkhorn clustering approach can be found in \cite{asano2019self,caron2020unsupervised}. 

\noindent\textbf{Prediction Target.} We investigate two sources of pseudo-labels as prediction targets. In the first approach, the \textbf{swap} prediction utilizes a pseudo-label created from a different domain as a prediction target. As shown in the yellow box of Figure \ref{fig:compare_cluster2} (c),  pseudo-labels from the audio (orange) and text (green) domains are used as prediction targets for the visual feature (blue). This mechanism is similar to XDC \cite{alwassel2020self} except that we perform this approach on projected features a in common space.
In the \textbf{joint} prediction method, a mean feature from the features of three modalities is first computed as a multimodal feature representation. Later, its pseudo-label will be the prediction target for the three separate feature instances and will be used to guide the features to be close across modalities and semantics. As shown in Figure \ref{fig:compare_cluster2} (d), the pseudo-label of the mean feature is used as the prediction target for features of each of the three modalities. 

\noindent\textbf{Label type.} We have two kinds of labels:  hard labels that represent discrete labels and soft labels that represent continuous, probabilistic labels. Since K-means assigns each instance to one of the centroids, it will only produce hard labels. The outputs from the Sinkhorn clustering are from a learnable network. We can use the softmax operator to transfer these outputs into probabilities over different labels (soft) or use the arg-max function to derive discrete labels (hard). When we perform soft-label prediction over the Sinkhorn pipeline as shown in (a), it will be similar to Swav \cite{caron2020unsupervised}, but we perform this over multiple modalities and treat the different modalities as a kind of data augmentation. 

\subsection{Dataset and computational resources used in each methods}
To better compare between different methods and settings, we specify various datasets used to construct each of the baselines in Tables \ref{tab:retrieval} and \ref{tab:temporal_state}. Methods with pre-trained feature extractors were trained on ImageNet (ImNet), Kinetics (K400), or Visual Genome (VG). Large-scale datasets such as HowTo100M (HT) and AudioSet (AS) are used for self-supervised pre-training. ActBERT \cite{zhu2020actbert} uses region features from a faster R-CNN, which is pre-trained on VG to better localize actions in CrossTask. We also include the computation resource and training time of each method. Note that methods \cite{alayrac2020self,miech2020end} with trainable backbones (TR) require 32 or more TPUs and usually perform better. For the reproduced *MIL-NCE method, we use code from \cite{miech2019howto100m} and apply the loss of \cite{miech2020end} from their Github repo. 

\begin{table}[t]
    \footnotesize
    \captionsetup{font=small}

    \centering
  \begin{tabular}{p{2cm} @{} c c c }
  \toprule
     Method &   MMS & MMS  + Clus & MMS  + Clus + Recon \\
    \midrule 
    \textbf{Aligned   $\uparrow$ }                        &  0.740 & 0.858 & 0.873\\
    \textbf{Misaligned  $\downarrow$ }             &   0.327 &    0.279 & 0.260 \\
    \bottomrule
  \end{tabular}
  \caption{ Cosine similarity of aligned and misaligned instances.}  \label{eq:cluster_result}
 \end{table}

\begin{table}[h]
		\tablestyle{2pt}{1.05}
		\resizebox{\columnwidth}{!}{
		\begin{tabular}{@{}l|c|ccc|ccc@{}}
			\toprule
			 \multicolumn{2}{c}{} & \multicolumn{3}{c}{YouCook2}                  & \multicolumn{3}{c}{MSRVTT}                   \\ 
			\cmidrule(lr){3-5} \cmidrule(lr){6-8} 
			Method        & Mod  & R@1  & R@5  & R@10 & R@1 & R@5  & R@10 \\ 
			\midrule
			MMS & T$\xrightarrow[]{}$V & 7.4 & 20.0 & 29.3 & 8.8& 23.2& 32.2 \\
			MIL-NCE*  & T$\xrightarrow[]{}$V & 8.1 & 23.3 & 32.3 & 8.4& 23.2& 32.4 \\
			Ours & T$\xrightarrow[]{}$V & 8.6 & 24.1 & 33.4 & 9.6& 23.4&  32.1 \\
			\midrule
			MIL-NCE* + audio & A$\xrightarrow[]{}$V & 16.2 &36.6 & 43.7 & 13.2& 28.4& 33.3 \\
			Ours & A$\xrightarrow[]{}$V & 19.4 & 41.3 & 50.9 & 14.8 & 30.1 & 39.0 \\
			\midrule
			NCE  & T$\xrightarrow[]{}$VA & 14.5 & 32.1 & 39.2 & 8.8& 24.1 & 33.7 \\
			MIL-NCE* + audio & T$\xrightarrow[]{}$VA & 15.1 & 31.9 & 40.0 & 9.0& 23.3 & 33.0 \\
			MMS  & T$\xrightarrow[]{}$VA & 16.1 & 33.9 & 43.7 & 9.5& 23.3 & 32.9 \\
			Ours & T$\xrightarrow[]{}$VA & 18.1 & 35.5 & 45.2 & 10.5& 25.2& 33.8 \\
			
			\bottomrule
			 
		\end{tabular}
		}
		\caption{Comparison of retrieval across different modalitites. 
		\label{tab:modality_ablation}
		}
		
\end{table}

\section{Additional experiments}
\subsection{Dealing with miss-alignment across modalities}
To quantify the alignment discrepancy across modalities, we first consider the pairwise MMS loss for each modality combination: AT, AV, and VT (V: video, A: audio, T: text). The loss starts equally for all combinations from (16.3, 16.8, 16.4) and decreases to AT=2.4, AV=8.8, VT=10.8 (epoch 10). The AT loss is the lowest since the text was generated from an ASR system, followed by AV since both signals are synchronized, which is relevant for object sounds like sizzling or chopping, and the largest gap can be found for VT pairs. Hence, introducing audio enables us to bridge this gap.
We hypothesize that the clustering loss implicitly compensates for this misalignment.
To show this effect, we sample V/A/T triplets from the YouCook2 dataset, generate misaligned instances by randomly replacing instances, and compare their cosine similarity to its mean multimodal embedding as in Eq.4 (see  Tab.~\ref{eq:cluster_result}, columns compare models from the ablation study). 
With the proposed clustering, aligned instances are closer to the mean embedding while misaligned are further away (as desired).
Therefore, the clustering step in training could compensate/correct for the MMS loss, which always pulls together true instances, even if they are misaligned.
With the proposed clustering, aligned instances are closer to the mean embedding while misaligned are further away, because the contrastive loss pulls every pair no matter the similarity between the instances. In the clustering step, for the aligned pairs, modalities will converge better while misaligned pairs will stay apart.

\subsection{Ablation of modalities.} 
We perform ablation experiments on the use of modalities in Table \ref{tab:modality_ablation}. From these experiments we find audio information to be crucial in bridging the gap between video and text while learning a joint space across the three modalities. The improvement on MSR-VTT is not significant compared to Youcook2. We attribute this performance difference to the domain gap between the various datasets. Both HowTo100M and Youcook are based on instructional videos where the text modality has a strong correlation to the video and audio modalities. In HowTo100M, the text is based on ASR transcripts. 
In Youcook2 and MSR-VTT, the query texts are hand-annotated captions. While Youcook2 captions describe single cooking steps, MSR-VTT captions are general descriptions of the scene, with captions. These captions are often not close to instructional ASR and also less related to what is being said in the audio.

\begin{table}[t]
\centering
\tablestyle{4pt}{1.05}
 		\begin{tabular}{@{}lcccc@{}}
 		    \toprule
 		     & \multicolumn{2}{c}{UCF-101}                  & \multicolumn{2}{c}{HMDB} \\
 			Method & Top-1 & Top-5 & Top-1 & Top-5 \\
 			\midrule
 			Brattoli \etal \cite{brattoli2020rethinking} & 37.6 & 62.5 & 26.9 & 49.8 \\
 			\midrule
 			MCN (ours) & 33.0 & 62.3 & 20.9 & 48.4 \\
 			MCN-actions (ours) & 33.9 & 63.7 & 22.5 & 51.5 \\
 			
 			\bottomrule
 		\end{tabular}
 		
 		\caption{Zero-shot action recognition performance on the UCF-101 and HMDB datasets. MCN-actions is the MCN method, which has been ``fine-tuned" on a subset of the HowTo100M dataset which contains action-related videos.
 		\label{tab:zs_actrec}
 		}
 \end{table} 
\begin{table}
		\tablestyle{2pt}{1.05}
		\resizebox{\columnwidth}{!}{
		\begin{tabular}{@{}l|c|cc|cccc@{}}
			\toprule
			 \multicolumn{4}{c}{} & \multicolumn{4}{c}{YouCook2} \\ 
			\cmidrule(lr){5-8} 
			Method        & Mod & Model  &  FT & R@1  & R@5  & R@10 & Median R \\ \midrule
			Random & & - & - & 0.03 & 0.15 & 0.3 & 1678 \\
			Miech  \cite{miech2019howto100m} & VT &  R152+RX101  & Y  & 8.2 & 24.5 & 35.3  & 24  \\
			\textbf{MCN (ours)} & VT & R152+RX101 & Y & 11.3 & 28.2 & 38.4  & 20 \\
            \textbf{MCN (ours)} & VAT & R152+RX101 & Y & \textbf{28.2} & \textbf{53.0} & \textbf{63.7}  & \textbf{5} \\
			\bottomrule
		\end{tabular}
		}
		\caption{Comparison of text-to-video retrieval systems on finetune setting. FT indicates if it is finetuned on the downstream dataset. 
		\label{tab:finetune}
		}
\end{table} 
\begin{table}
		\tablestyle{2pt}{1.05}
		\resizebox{\columnwidth}{!}{
		\begin{tabular}{@{}c|cccc|cccc@{}}
			\toprule
			 \multicolumn{1}{c}{} & \multicolumn{4}{c}{YouCook2}                  & \multicolumn{4}{c}{MSRVTT}                   \\ 
			\cmidrule(lr){2-5} \cmidrule(lr){6-9} 
			Cluster size $k$         & R@1  & R@5  & R@10 & Median R & R@1 & R@5  & R@10 & Median R \\ \midrule
			64   & 17.8 & 34.7 & 43.4   & 17 & 10.1 & 25.3 & 34.1 & 27 \\
			128   & 17.3 & 34.8 & 44.2   & 19 & \textbf{10.5} & 24.5 & 33.5 & 29 \\
			256   & 18.1 & \textbf{35.5} & \textbf{45.2}   & \textbf{16} & \textbf{10.5} & 25.2 & 33.8 & 27 \\
			512   & \textbf{18.3} & 35.3 & 44.4   & 19 & 10.4 & 24.6 & 33.5 & 26.5 \\
			1024   & 17.9 & 34.6 & 43.5   & 17 & 9.4 & \textbf{25.8} & \textbf{34.6} & \textbf{25} \\
			\bottomrule
		\end{tabular}
		}
		\caption{Comparison of text-to-video retrieval systems on different number of cluster size in K-means 
		\label{tab:cluster}
		}
\end{table}

\begin{table*}
		\tablestyle{2pt}{1.05}
		\resizebox{2\columnwidth}{!}{
		\begin{tabular}{@{}l|c|ccccc|ccc|ccc@{}}
			\toprule
			 \multicolumn{7}{c}{} & \multicolumn{3}{c}{YouCook2}                  & \multicolumn{3}{c}{MSRVTT}   \\ 
			\cmidrule(lr){8-10} \cmidrule(lr){11-13} 
Method        & Mod & Model & Dataset & Com & Time & TR & R@1  & R@5  & R@10 & R@1 & R@5  & R@10 \\ \midrule
Random & & - & & - & & & 0.03 & 0.15 & 0.3 & 0.01& 0.05& 0.1 \\
Miech  \cite{miech2019howto100m} & VT &  R152+RX101 & HT+ImNet+K400 & 1 V100 & 1 day & N  & 6.1 & 17.3 & 24.8                        & 7.2 & 19.2 & 28.0      \\
MDR \cite{amrani2020noise} & VT & R152+RX101 & HT+ImNet+K400 &1 V100& 1 day& N & - & - &- & 8.0 & 21.3 & 29.3 \\
MIL-NCE* \cite{miech2020end}& VT  & R152+RX101 & HT+ImNet+K400& 4 V100 &2 days& N  & 8.1 & 23.3 & 32.3                        & 8.4 & 23.2 & 32.4                      \\
\textbf{MCN (ours)} & VAT & R152+RX101 & HT+ImNet+K400 & 4 V100 &2 days& N & \textbf{18.1} & 35.5 & 45.2                        & \textbf{10.5} & \textbf{25.2} & \textbf{33.8}                        \\
\midrule
MDR \cite{amrani2020noise} &VT & R152 & HT+ImNet+K400 &1 V100&1 day& N & - & - &- & 8.4 & 22.0 & 30.4 \\
ActBERT \cite{zhu2020actbert} &VT & R101+Res3D & HT+VG+K400 &&& N  & 9.6 & 26.7 & 38.0                        & 8.6 & 23.4 & 33.1                      \\
SSB \cite{patrick2020support}& VT  & R(2+1)D-34+R152&  HT  & 8 V100&1 day& N  & - & - & -                        & 8.7 & 23.0 & 31.1                      \\

\midrule
MMV FAC  \cite{alayrac2020self}          &VAT & TSM-50x2 & HT+AS& 32 TPU &3 days& Y   & 11.7 & 33.4 & 45.4             & 9.3 & 23.0 & 31.1                      \\
MIL-NCE~\cite{miech2020end} &VT & I3D-G   &  HT& 64 TPU&3 days& Y & 11.4 & 30.6 & 42.0  & 9.4 & 22.0 & 30.0                    \\
MIL-NCE~\cite{miech2020end} &VT & S3D-G   & HT & 64 TPU &3 days& Y & 15.1 & \textbf{38.0} & \textbf{51.2}  & 9.9 & 24.0 & 32.4                    \\

\bottomrule
			 
		\end{tabular}
		}
		\caption{Comparison of text-to-video retrieval systems. Mod indicates modality used, where V: video, A: audio, T: text. HT: HowTo100M. VG: Visual Genome. AS: AudioSet. Com stands for computational resource. Time indicates the training time. TR indicates if a trainable backbone is used or not. 
		\label{tab:retrieval}
		}
\end{table*} 

\begin{table*}
		\tablestyle{2pt}{1.05}
		\resizebox{2\columnwidth}{!}{
		\begin{tabular}{@{}l|c|ccccc|ccc|ccc@{}}
			\toprule
			 \multicolumn{7}{c}{}& \multicolumn{3}{c}{CrossTask} & \multicolumn{3}{c}{MYT}   \\ 
			\cmidrule(lr){8-10} \cmidrule(lr){11-13} 
			Method  & Mod   & Model & Dataset & Com & Time &TR   & Recall & IOD & IOU & Recall & IOD & IOU \\ 
			\midrule
			CrossTask \cite{zhukov2019crosstask}  &VT & R152+I3D & CrossTask &&& N  & 22.4 & - & -  & -  & - & -  \\
			CrossTask \cite{zhukov2019crosstask}  &VT & R152+I3D & CrossTask &&& N  & 31.6  & - & - & -  & - & -  \\
			Mining: GRU \cite{kuehne2019mining}  &VT  & TSN & MiningYouTube &&& N  & -  & - & -  &   - & 14.5 & 7.8  \\
			Mining: MLP \cite{kuehne2019mining}  &VT & TSN & MiningYouTube &&& N   & -  & - & -   & - & 19.2 & 9.8  \\
			\midrule
			Miech  \cite{miech2019howto100m}  &VT  & R152+RX101 & HT+ImNet+K400 &1 V100&1 day& N   & 33.6 & 26.6 & 17.5   & 15.0 & 17.2 & 11.4 \\
			MIL-NCE* \cite{miech2020end} &VT & R152+RX101& HT+ImNet+K400 &4 V100&2 days& N & 33.2 & 30.2 & 16.3   & 14.9 & 26.4 & 17.8                      \\
            
			\textbf{MCN (ours)}  &VAT & R152+RX101 &HT+ImNet+K400&4 V100& 2 days & N & 35.1 & \textbf{33.6} & \textbf{22.2}                        & \textbf{18.1}& \textbf{32.0} & \textbf{23.1}                        \\
			\midrule
			ActBERT \cite{zhu2020actbert}  &VT & R101+Res3D &HT+K400&&& N & 37.1 & - & -   &  - & - & - 
			\\
			ActBERT \cite{zhu2020actbert}  &VT & + Faster R-CNN &HT+VG+K400&&& N & \textbf{41.4} & - & -   &  - & - & - 
			\\
			\midrule

			  MIL-NCE \cite{miech2020end}  &VT & I3D-G& HT &64 TPU&3 days& Y & 36.4 & - & -   & - & - & -                      \\
			  MIL-NCE \cite{miech2020end}  &VT & S3D-G & HT &64 TPU&3 days& Y & 40.5 & - & -   & - & - & -                      \\
			
			\bottomrule
		\end{tabular}
		}
		\caption{Evaluation of temporal action localization systems.  
		\label{tab:temporal_state}
		}
\end{table*}

\subsection{Zero-Shot Action Recognition} 
We also test our method's performance for the downstream task of zero-shot action recognition. For these experiments, we follow the evaluation protocol of \cite{brattoli2020rethinking} and test on the full UCF-101 and HMDB datasets. We present the top-1 and top-5 accuracies on both datasets in Table \ref{tab:zs_actrec}. Although MCN is trained using instructional videos, we find that the joint video/text space it learns is sufficient for the task of zero-shot action recognition. Furthermore, our method can be further improved by training on action-related videos; by removing various video categories - 'food and entertaining', 'computers and electronics', 'cars and other vehicles', 'home and garden', and 'health' and training on a subset of the HowTo100M dataset, we find MCN is able to achieve state-of-the-art Top-5 accuracy on both datasets. The baseline, \cite{brattoli2020rethinking}, is a method designed specifically for zero-shot action recognition and is trained using labeled action videos from Kinetics-700, leading to strong top-1 accuracy. 

\begin{table*}[t!]
\resizebox{\textwidth}{!}{
\begin{tabular}{lc@{~~~~}c@{~~}c@{~~}c@{~~}c@{~~}c@{~~}c@{~~}c@{~~}c@{~~}c@{~~}c@{~~}c@{~~}c@{~~}c@{~~}c@{~~}c@{~~}c@{~~}c@{~~}|c} \toprule
    & \rotatebox{90}{\small Make} \rotatebox{90}{\small Kimchi Rice}  
    & \rotatebox{90}{\small Pickle} \rotatebox{90}{\small Cucumber}  
    & \rotatebox{90}{\small Make Banana} \rotatebox{90}{\small Ice Cream}  
    & \rotatebox{90}{\small Grill} \rotatebox{90}{\small Steak}  
    & \rotatebox{90}{\small Jack Up } \rotatebox{90}{\small Car}  
    & \rotatebox{90}{\small Make } \rotatebox{90}{\small Jello Shots}  
    & \rotatebox{90}{\small Change } \rotatebox{90}{\small Tire}  
    & \rotatebox{90}{\small Make } \rotatebox{90}{\small Lemonade}  
    & \rotatebox{90}{\small Add Oil } \rotatebox{90}{\small to Car}  
    & \rotatebox{90}{\small Make } \rotatebox{90}{\small Latte}  
    & \rotatebox{90}{\small Build } \rotatebox{90}{\small Shelves}  
    & \rotatebox{90}{\small Make } \rotatebox{90}{\small Taco Salad}  
    & \rotatebox{90}{\small Make } \rotatebox{90}{\small French Toast}  
    & \rotatebox{90}{\small Make } \rotatebox{90}{\small Irish Coffee}  
    & \rotatebox{90}{\small Make } \rotatebox{90}{\small Strawberry Cake}  
    & \rotatebox{90}{\small Make } \rotatebox{90}{\small Pancakes}  
    & \rotatebox{90}{\small Make } \rotatebox{90}{\small Meringue}  
    & \rotatebox{90}{\small Make } \rotatebox{90}{\small Fish Curry}  
    & \rotatebox{90}{\small Average }
\\ \midrule

CrossTask \cite{zhukov2019crosstask}                         & 13.3          & 18.0          & 23.4 & 23.1 & 16.9          & 16.5 & 30.7 & 21.6 & 4.6          & 19.5 & 35.3        & 10.0 & 32.3 & 13.8 & 29.5 & 37.6 & {43.0} & 13.3 & 22.4 \\ 
Supervised \cite{zhukov2019crosstask}                           & 19.1          & 25.3          & 38.0          & 37.5          & 25.7          & 28.2          & \textbf{54.3}          & 25.8          & 18.3         & 31.2          & \textbf{47.7}          & 12.0          & \textbf{39.5}          & 23.4          & 30.9          & 41.1          & \textbf{53.4}          & 17.3          & 31.6 \\ 
Miech \etal ~\cite{miech2019howto100m}                    & \textbf{33.5}           & {27.1}         & {36.6}           & \textbf{37.9}           & {24.1}           & \textbf{35.6} & {32.7}           & {35.1}           & \textbf{30.7}         & {28.5}           & {43.2}         & {19.8}           & {34.7}           & {33.6}           & \textbf{40.4}           & {41.6}          & 41.9           & {27.4}           & {33.6} \\
\midrule 

MCN & 25.5 &    \textbf{31.1} &    \textbf{39.7} &    32.7 &    \textbf{35.4} &    36.8 &    29.0 &    \textbf{40.0} &    28.4 &    \textbf{33.8} &    45.7 &    \textbf{27.5} &    36.1 &    \textbf{34.9} &    39.6 &    \textbf{42.6} &    43.0 &    \textbf{29.1} & \textbf{35.1} \\
\bottomrule
\end{tabular}
}
\caption{\textbf{Action step localization} results on CrossTask.}
\label{table:action_step_localization}
\end{table*}

\subsection{CrossTask specific results.}
We break down the consolidated performance result reported in the main paper on CrossTask and show results corresponding to each specific task in Table \ref{table:action_step_localization}. We observe that our model shows a very different yet often improved performance pattern, compared to the visual-only features used in \cite{miech2019howto100m} and \cite{zhukov2019crosstask}. We attribute this behavior to varying levels of information provided by the audio modality in each setting.
\subsection{Finetune results} 
We show our model's performance on the finetune setting in Table \ref{tab:finetune}, which means we also train on an additional training set provided by the Youcook \cite{zhou18towards} dataset. Although the finetune setting, which requires ground-truth labels, isn't our main focus, we obtain significant improvement over the current baseline.
\subsection{Different number of clusters}
Table \ref{tab:cluster} shows the results using different number of cluster sizes for K-means. The result shows similar performance across different datasets and evaluation metrics.
%

%
%
%
%
%
%
%
%
%
%
\subsection{Audio length used in inference.}
\begin{figure}
 \centering
  \includegraphics[width=0.9\columnwidth]{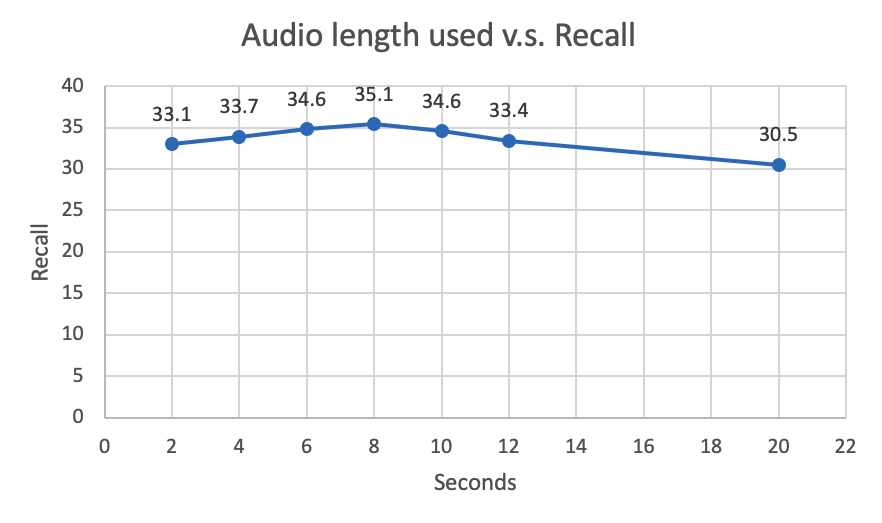}
\caption{Audio length used in inference on CrossTask.}
\label{fig:audio_length}
\end{figure}

We test the audio length needed for effective inference performance on CrossTask. As shown in Fig \ref{fig:audio_length}, we find that using 8 seconds (4 seconds before and after) of audio leads to the best results. Given that some steps are very short (less than 3 seconds), this result also shows that using very long audio segments can distract the model from predicting a correct localization step.

\begin{figure*}[tbhp]
 \centering
  \includegraphics[width=2\columnwidth]{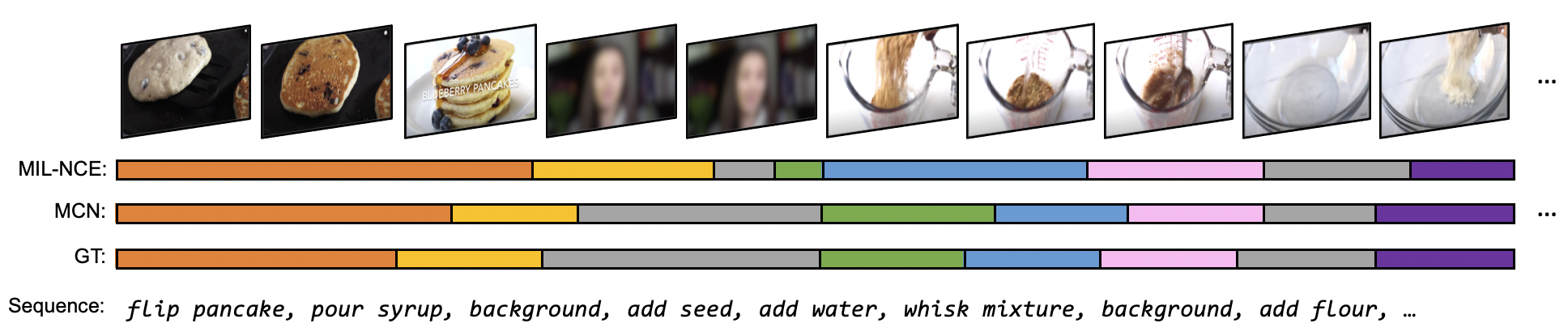}
\caption{Temporal action localization example from the first minute of the video "Vegan Blueberry Quinoa Pancakes" in the MiningYouTube dataset. Given the video and the action step sequence, the goal is to align the step temporal boundaries.}
\label{fig:action_ex}
\end{figure*}

 \begin{figure*}[h!]
 \centering
  \includegraphics[width=2\columnwidth]{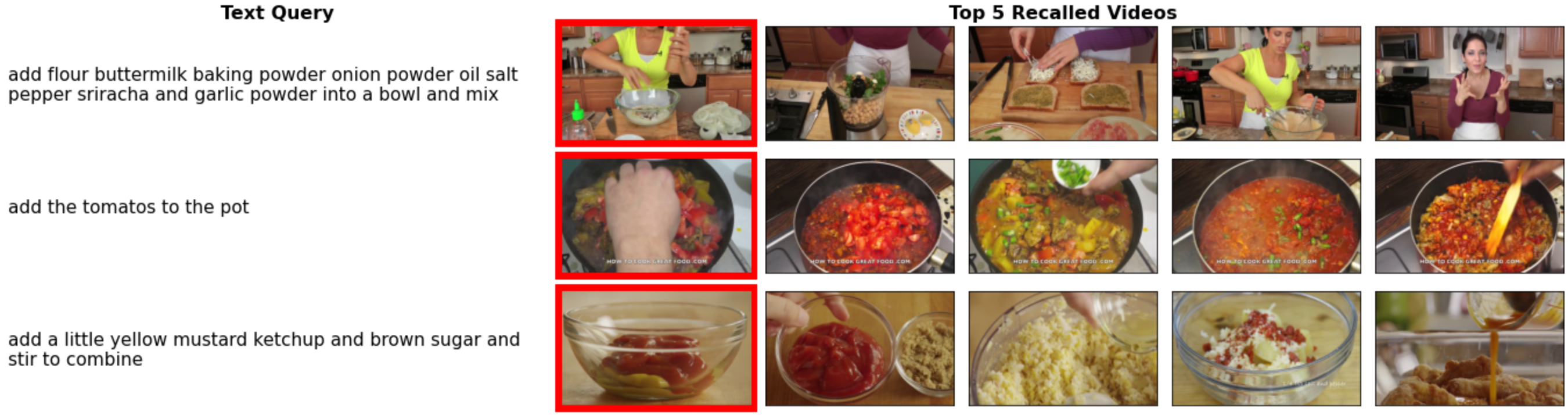}
\caption{Text-to-video retrieval examples. The retrieved video clips show a similar pattern.}
\label{fig:retrieve_ex}
\end{figure*}

%
%

\section{Qualitative analysis}
To further understand the proposed MCN model's improved performance, we also perform a qualitative analysis with the model's temporal action localization results on the MiningYoutube task. One interesting observation is shown in Figure \ref{fig:action_ex}. We observed that our model performs well in distinguishing action steps from the background scenes. We attribute this improvement to the proposed clustering component, which we observe has separated the background frames from various action classes. Background class instances are often placed as outliers with respect to the various action step clusters.
In Figure \ref{fig:retrieve_ex}, we show more examples on text-to-video retrieval. The retrieved video segments show similar semantics.
%
{\small
\bibliographystyle{ieee_fullname}
\bibliography{MCN}
}

\end{document}